\documentclass[acmtog,screen,nonacm]{acmart}

\AtBeginDocument{%
  }

\usepackage{soul}
\usepackage{amsmath}
\usepackage{amssymb}
\usepackage{microtype}
\usepackage{wrapfig}
\usepackage{fancyhdr}
\usepackage{mathtools}
\usepackage{xspace}
\usepackage{float}
\usepackage{colortbl} % enables \cellcolor in tables
\usepackage{xcolor}    % options already set in \documentclass
\usepackage{bm}
\usepackage{multirow}
\usepackage{graphicx}  % <-- needed for images and DeclareGraphicsExtensions
\DeclareGraphicsExtensions{.png,.jpg,.pdf} % safe extensions

\def\figurePath{figures/}

\def\myfigure#1#2#3{\begin{figure}[t]\centering\includegraphics*[width = #3\columnwidth]{\figurePath#1}\\[-2.5ex]\caption{#2}\label{fig:#1}\end{figure}}    

\def\mycfigure#1#2{\begin{figure*}[t]\centering\includegraphics*[clip, width = \textwidth]{\figurePath#1}\\[-1.5ex]\caption{#2}\label{fig:#1}\end{figure*}}

\def\myccfigure#1#2#3{\begin{figure*}[t]\centering\includegraphics*[clip, width = #3\textwidth]{\figurePath#1}\\[-1.5ex]\caption{#2}\label{fig:#1}\end{figure*}}

\newcommand{\refSec}[1]{Sec.~\ref{sec:#1}}
\newcommand{\refFig}[1]{Fig.~\ref{fig:#1}}
\newcommand{\refEq}[1]{Eq.~\ref{eq:#1}}
\newcommand{\refTbl}[1]{Tbl.~\ref{tab:#1}}

\soulregister\ref7
\soulregister\cite7
\soulregister\refFig7
\soulregister\refSec7
\soulregister\ref7
\soulregister\pageref7
\soulregister\shortcite7
\soulregister\eg0
\soulregister\ie0

\DeclareGraphicsExtensions{.png,.jpg,.pdf,.ai,.psd}
\newcommand{\colorIcon}[2]{\textcolor{color#1}{\csname icon#2\endcsname}}

\newcommand{\nameAndIcon}[1]{``#1'' ({\csname icon#1\endcsname})}
\newcommand{\nameAndIconA}[1]{(``#1'', {\csname icon#1\endcsname})}
\begin{document}
% \title{Learning Reliable Full-Reference Image Quality Assessment via Multi-Scale CNNs and Metric-Ensemble Supervision}
\title{MILO: A Lightweight Perceptual Quality Metric for Image and Latent-Space Optimization}
%MILO: A Lightweight Multi-Scale Metric for Image and Latent-Space Perceptual Optimization
% MILO = Perceptual Reference-based Image Quality Scoring and Masking
% MILO: A Lightweight Perceptual Metric for Image and Latent-Space Optimization

\begin{teaserfigure}
  \includegraphics[width=\textwidth]{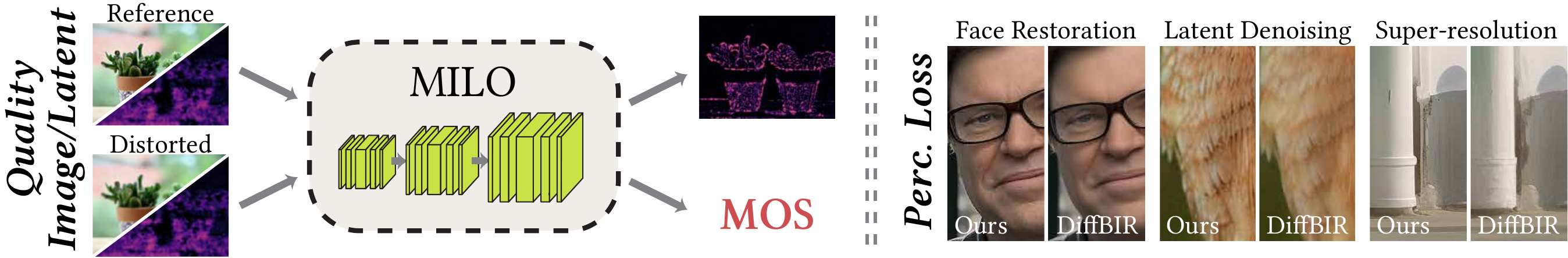}
  \caption{We present \textit{MILO}, our perceptual metric for full-reference image quality assessment, \textbf{solely trained on learned perceptual data}. MILO processes either RGB or VAE-decoded latent input pairs and outputs a perceptual quality score in the MOS range along with a spatial visibility map. When applied as a perceptual loss function, MILO allows through a curriculum learning strategy for spatially guided and content-aware refinement in image restoration tasks and diffusion-based optimization.}
  \label{fig:teaser}
\end{teaserfigure}


%% ORCID
% Ugur  \orcid{0000-0003-1279-6918} 
% M. Bemana \orcid{0000-0003-0202-6266} 
% HP. Seidel \orcid{0000-0002-1343-8613} 
% K. Myszkowski \orcid{0000-0002-8505-4141}
% Colin \orcid{0000-0001-6445-5563}

\author{U\u{g}ur \c{C}o\u{g}alan}
\affiliation{%
  \institution{Max Planck Institute for Informatics}
  \city{Saarbrücken}
  \country{Germany}}
\email{ugurcogalan@gmail.com}

\author{Mojtaba Bemana}
\affiliation{%
  \institution{Max Planck Institute for Informatics}
  \city{Saarbrücken}
  \country{Germany}}
\email{mbemana@mpi-inf.mpg.de}  

\author{Karol Myszkowski}
\affiliation{%
  \institution{Max Planck Institute for Informatics}
  \city{Saarbrücken}
  \country{Germany}}
\email{karol@mpi-inf.mpg.de}  

\author{Hans-Peter Seidel}
\affiliation{%
  \institution{Max Planck Institute for Informatics}
  \city{Saarbrücken}
  \country{Germany}}
\email{hpseidel@mpi-inf.mpg.de}  

\author{Colin Groth}
\affiliation{%
  \institution{Max Planck Institute for Informatics}
  \city{Saarbrücken}
  \country{Germany}}
\email{c.groth@nyu.edu}

\renewcommand{\shortauthors}{\c{C}o\u{g}alan et al.}

\begin{abstract}

We present MILO (Metric for Image- and Latent-space Optimization), a lightweight, multiscale, perceptual metric for full-reference image quality assessment (FR-IQA).
MILO is trained using pseudo-MOS (Mean Opinion Score) supervision, in which reproducible distortions are applied to diverse images and scored via an ensemble of recent quality metrics that account for visual masking effects.
This approach enables accurate learning without requiring large-scale human-labeled datasets. Despite its compact architecture, MILO outperforms existing metrics across standard FR-IQA benchmarks and offers fast inference suitable for real-time applications.
Beyond quality prediction, we demonstrate the utility of MILO as a perceptual loss in both image and latent domains. In particular, we show that spatial masking modeled by MILO, when applied to latent representations from a VAE encoder within Stable Diffusion, enables efficient and perceptually aligned optimization. By combining spatial masking with a curriculum learning strategy, we first process perceptually less relevant regions before progressively shifting the optimization to more visually distorted areas. This strategy leads to significantly improved performance in tasks like denoising, super-resolution, and face restoration, while also reducing computational overhead.
MILO thus functions as both a state-of-the-art image quality metric and as a practical tool for perceptual optimization in generative pipelines.

\end{abstract}

% \cg{
% Notes:
% - need for perceptual awareness in latent space
% - highlighting of the advantages of operating in latent space: computational speed, access to semantic features, etc
% - general-purpose perceptual tool not only good IQA metric -> seamlessly integrates with modern generative pipelines.
% } 



\settopmatter{printacmref=false}
\setcopyright{none}
\renewcommand\footnotetextcopyrightpermission[1]{}
\pagestyle{plain}


\maketitle

\section{Introduction}
\label{sec:intro}

% Visual content has long shaped how humans capture and interpret the world around them.
%For centuries, people have captured significant moments through photographs and expressed imaginary scenes in paintings. Today, this visual expression increasingly relies on computer-generated imagery, with a fast-growing focus on entirely synthesized images by deep neural networks. 
%The demand for high-quality image generation drives an explosion in image generation networks that strive to create visuals that are both technically accurate and perceptually aligned with human vision.
%At the core of advancing these models lies the need for reliable metrics---tools that can assess how closely generated images align with visual expectations. In image synthesis and restoration pipelines, these metrics evaluate the visual fidelity of generated content and serve as optimization targets. 

For centuries, people have captured real and imagined scenes through photography and painting. Today, visual expression increasingly relies on computer-generated imagery, especially images synthesized by deep neural networks. As demand for high-quality generation grows, so does the need for reliable metrics to assess visual quality as perceived by the human observer and guide optimization in image synthesis and restoration pipelines.

%Traditional metrics, such as Mean Absolute Error (MAE) and Peak Signal-to-Noise Ratio (PSNR), rely on per-pixel comparisons. While these methods often fail to capture accurate human perception, resulting in images that may be statistically correct but visually unsatisfying.
%In response, perceptually motivated metrics like LPIPS \cite{zhang2018perceptual} and DISTS \cite{ding20} have emerged, utilizing deep features from pre-trained classification networks, e.g., VGG, to align more closely with human perception. 
%Other recent learnable metrics are directly trained on perceptual data to capture subjective quality judgments from human feedback \cite{prashnani2018pieapp, cougalan2024enhancing, bosse2017deep}. 
%While training those state-of-the-art metrics on perceptual datasets proved to create the most visually pleasing results, such datasets come with unique challenges. 

Traditional metrics like Mean Absolute Error (MAE) and Peak Signal-to-Noise Ratio (PSNR) rely on per-pixel comparisons and often fail to align with human perception. Perceptual metrics such as LPIPS \cite{zhang2018perceptual} and DISTS \cite{ding20} improve on this by leveraging deep features from pre-trained networks like VGG. These methods are trained on human-labeled perceptual data at the patch level, which is costly to collect and may still fall short of capturing perceived quality in the context of full images.

Due to the labor-intensive nature of repeated and controlled human judgment collection, perceptual datasets are limited in scope. For example, the widely used KADID-10k dataset \cite{lin19} contains only 81 reference images, each with 25 distortions at five distortion levels, amounting to over 10,000 image pairs that need annotation. 
% As a result, learning-based IQA models trained on such data often generalize poorly to unseen content or distortion types. 
This scarcity is particularly problematic given that perceived quality is strongly dependent on image content and local context. As illustrated in Fig.~\ref{fig:same_MOS2}, even large variations in distortion strength can result in identical subjective ratings due to masking or content-based tolerance. 
% Another problem with current perceptual datasets is the limited number of different types of distortions and the reproducibility of such distortions, especially for ML-based distortion techniques, where the missing forward model makes reproduction challenging.
However, the underlying challenge extends beyond data scarcity: it lies in the absence of models that capture variations in perceptual sensitivity across spatial structure and frequency content. Effective IQA methods must account for these factors to predict human judgments reliably and support optimization tasks that require fine-grained, spatially aware perceptual guidance.

We propose MILO, a compact and efficient multiscale FR-IQA metric that addresses these limitations. MILO is trained using pseudo-MOS labels generated by an ensemble of visual masking-aware metrics applied to synthetically distorted images through a forward model. This strategy enables large-scale supervision without the need for manual annotations. The resulting model predicts both global perceptual scores and spatial visibility maps, which allow for the localization of the distortion visibility based on the underlying content. MILO achieves state-of-the-art performance on common IQA benchmarks while being significantly faster than existing learning-based metrics due to its lightweight structure.
Beyond quality prediction, we show that MILO is effective as a perceptual loss in restoration and generative tasks. In particular, our method can operate directly in the latent space of diffusion models. Most contemporary generative pipelines perform image synthesis through iterative refinement of the latent. Applying perceptual guidance in this domain is advantageous, as it avoids repeated decoding into image space and allows perceptual supervision to act on high-level semantic representations. Our method enables spatially resolved loss weighting in latent space, improving the quality and efficiency of diffusion-based optimization.
% Our findings suggest that MILO is not only a fast and accurate FR metric but also a practical tool for perceptual optimization in modern generative pipelines. 
The code for our metric, trained models, and further information are available at \url{https://milo.mpi-inf.mpg.de/}.
%
% In this work, we address the limitations of current perceptual quality datasets and metrics by introducing a two-part framework that improves both data diversity and metric performance in full-reference image quality assessment (FR-IQA). First, we propose a data augmentation strategy for FR perceptual datasets. By applying reproducible distortion models—such as Gaussian blur and additive noise—to arbitrary reference images, we generate new reference-distortion pairs. To assign reliable perceptual scores to these augmented samples, we leverage an ensemble of existing quality metrics. We show that this ensemble approach provides more accurate and robust predictions than any individual metric, enabling the creation of high-confidence training data without requiring additional human annotations.
% Second, we present a learnable FR-IQA metric based on a low-dimensional, multi-layer convolutional neural network. The architecture is purposefully lightweight to ensure fast runtime performance and practical deployment, including use as a perceptual loss during optimization. Despite its compact design, the model incorporates multiscale processing and achieves state-of-the-art accuracy. The metric not only predicts a perceptual quality score but also generates a visibility map, offering interpretability by localizing perceptual degradation.
%
Our contributions summarize as:
%
% In summary, our contributions are as follows:
%
% \vspace{-0.5cm}
\begin{itemize}
\item a fast perceptual metric (MILO) that achieves state-of-the-art accuracy in full-reference image quality assessment.
\item a pseudo-supervision strategy based on reproducible distortions and a metric ensemble to synthesize large-scale, annotated training data.
\item extension of our metric to operate directly in the latent space of diffusion models for efficient, perceptual optimization.
\item an effective perceptual loss in restoration tasks through curriculum learning and spatial masking.
\end{itemize}

\myfigure{same_MOS2}{Three distortion images of the KADID10k dataset that apply significantly different amounts of noise to the image. Yet, the subjective quality (MOS) is rated the same by human observers.}{1.0}
\section{Related work} \label{sec:backgound}
This section explores former research and discusses advancements in image quality assessment and data augmentation techniques.

\subsection{Image Quality Assessment} Image quality assessment (IQA) is primarily divided into full-reference (FR) and no-reference (NR) approaches. In this paper, we mainly focus on FR metrics. Traditional FR-IQA methods, such as MAE, PSNR, and SSIM \cite{wang04}, rely on direct comparisons between reference and distorted images to evaluate quality. 
FSIM \cite{zhang11} incorporates phase congruency and gradient magnitude to better model human perception. FLIP \cite{andersson2020flip} is a task-specific metric designed to assess perceptual differences in rendered graphics with high spatial precision.
However, these traditional metrics often fall short of representing visual quality as perceived by humans because they do not account for image content, e.g., high-frequency vs. low-frequency background, or perceptual phenomena such as masking.
In recent years, learning-based models \cite{prashnani2018pieapp,zhang2018perceptual,ding20,ding21,liao22} have been developed to more accurately predict perceptual quality, leveraging deep image features to better align with human perception. Notably, \citet{zhang2018perceptual} demonstrated that internal image representations from classification networks could be used to assess image similarity. Their Perceptual Image Patch Similarity (LPIPS) index computes image similarity by measuring $\ell_2$ distances between features extracted from pre-trained VGG networks. This idea builds on earlier perceptual feature-based losses, such as those used in style transfer and super-resolution networks \cite{johnson2016perceptual}, where VGG features provided a proxy for perceptual content.
%Ding et al. 
\citet{ding20} introduced the DISTS metric, which evaluates texture and structure similarity by comparing the global statistics—mean, variance, and correlations—of feature pairs, similar to SSIM. 
Shifting away from deterministic feature comparisons, DeepWSD \cite{liao22} employed the Wasserstein distance, a statistical method for comparing the distributions of features, offering a more holistic approach to IQA. 
TOPIQ \cite{chen2024topiq} extends this idea further by leveraging token-based representations from transformers, enabling better generalization across image types.
More recently, %Çoğalan et al. 
\citet{cougalan2024enhancing} proposed enhancing FR-IQA by learning visual masking models that modulate the metric input in a self-supervised manner using MOS data. Their masking network improves both traditional and learning-based metrics by integrating perceptual phenomena such as masking—an aspect also explicitly modeled in perceptually grounded methods like HDR-VDP-2 \cite{mantiuk2011hdr} and FovVideoVDP \cite{mantiuk2021fovvideovdp}, which incorporate display parameters and foveation but remain limited in general applicability due to their complexity and reliance on calibrated inputs.
Despite these advances, many of these methods are trained on relatively small datasets, such as KADID \cite{lin19} and PIPAL \cite{gu2020pipal}, limiting their ability to generalize to larger, more varied image collections. For example, WaDIQaM \cite{bosse2017deep} demonstrated promising performance by combining deep feature regression with patch-wise quality estimation, but it still depends on limited per-image labeled data.
However, the reliance on manually annotated datasets ---which require extensive human labor--- restricts both the size and diversity of training samples. 
In this paper, we propose a learnable FR-IQA metric that addresses these limitations by leveraging an augmented dataset with automatically assigned perceptual scores and a lightweight CNN architecture capable of generalizing across diverse image content. Compared to existing perceptually optimized metrics, our method aims to generalize across both natural and synthetic distortions.

\subsection{Data Augmentation}
\label{sec:DataAugment}
Human-labeled datasets are essential for training perceptually aligned IQA models, as demonstrated by PieApp, WaDIQaM, and the work of \citet{cougalan2024enhancing}. However, creating such datasets is costly and time-consuming, and their limited size restricts generalization across diverse content and distortions. 
% Additionally, in full-reference IQA, the perceptual relationship between the reference and its distorted counterpart must be preserved. 
This makes conventional augmentation strategies—common in classification or segmentation—largely inapplicable.
Patch sampling \cite{ahmed2022deep, bosse2017deep} and image mixing \cite{shi24dsmix} have been used to increase data extent. DSMix blends two distorted images with a binary mask, assuming a linear perceptual relationship between them—an untested and likely unreliable premise for human quality judgments. Similarly, random geometric or photometric transformations \cite{chen2024promptiqa} aim to improve metric robustness but do not introduce new perceptual content. 
% These strategies are fundamentally limited by their dependence on recombining existing samples rather than synthesizing perceptually novel data.

Recent work explores the use of multimodal large language models (MLLMs) for text-driven IQA augmentation \cite{you2023depicting, wu2024comprehensive}. \citet{wu2024comprehensive} evaluated various prompting techniques and psychophysical procedures to be used for IQA, finding that current MLLMs fail to reliably model fine-grained perceptual differences, such as subtle color shifts or quality distinctions between similar images.
Diffusion models are also promising, capable of generating diverse and semantically consistent images by learning probabilistic distributions \cite{ho2020denoising, saharia2022image}. They have been applied successfully in domains like medical imaging and remote sensing \cite{ozbey2023restoring}. Methods like DatasetDM \cite{wu2023datasetdm} enable perception-aware augmentation with corresponding annotations. However, these models still struggle with generating images suitable for reliable quality assessment.

In contrast, we propose a distortion-aware augmentation strategy based on reproducible forward models, generating perceptually diverse reference-distortion pairs. Scores from an ensemble of existing metrics provide pseudo-supervision, eliminating the need for human annotations while ensuring perceptual validity. This strategy allows for massive training of our learning-based FR metric.

\section{Metric for Image- and Latent-Space Optimization (MILO)}
\mycfigure{pipeline5}{Overview of the MILO pipeline. The method takes a pair of input images (or their VAE-encoded latents)---reference $X$ and distorted $Y$---and processes them at multiple scales. At each scale, a shared CNN module predicts a residual visibility mask $M_r$, which is recursively refined across scales. The final visibility map $M_V$ highlights perceptually salient distortion regions. A complementary global quality score in the MOS scale reflects predicted human judgment of perceptual quality.}{}
We propose MILO, a perceptual quality metric that predicts both a global quality score and a spatial visibility map for a given reference-distorted image pair.
However, as discussed in \refSec{DataAugment} existing perceptual datasets are limited in diversity and scale, restricting generalization---particularly for content-sensitive distortions. %For instance, varying noise intensity for smooth and textured regions may be perceived the same (see \refFig{same_MOS2}).

To address this limitation, we introduce a two-stage framework. First, we generate a synthetic training set by applying parameterized distortions (e.g., Gaussian blur, additive noise) to diverse images (see \refSec{mosPredict}). Corresponding perceptual scores are assigned via an ensemble of masking-aware full-reference metrics \cite{cougalan2024enhancing}, yielding dense and content-aware supervision without manual annotations.
Second, we train a lightweight multi-scale CNN that predicts both MOS and a pixel-wise visibility map. The architecture (see \refFig{pipeline5}) combines residual blocks and multiscale features to model content-dependent distortion visibility efficiently (see \refSec{multiCNN}). Our metric generalizes beyond its pseudo-supervision, producing state-of-the-art performance on standard IQA benchmarks.

Our metric is further extended to operate directly on VAE encoded latents to avoid costly image-space decoding (see \refSec{latentMasking}) and employ curriculum learning to schedule loss weighting from low-visibility to high-visibility regions for perceptually aligned optimization (see \refSec{curriculum}).

\subsection{Training Data Augmentation}\label{sec:mosPredict}
To augment existing IQA datasets with reliable perceptual annotations, we develop a pseudo-MOS prediction module based on an ensemble of masking-aware FR-IQAs. The goal is to assign perceptual scores to newly generated distorted images without relying on extensive studies with human participants.
We leverage the forward distortion model provided by the KADID10k dataset \cite{lin19} to synthetically reproduce traditional distortion types (e.g., Gaussian blur, additive noise, JPEG compression) at controlled levels on arbitrary reference images. This procedure ensures that each generated image follows a known and interpretable degradation process, instead of handling learning-incurred distortions or domain-specific artifacts. The reference images for the data augmentation are randomly chosen from the ImageNet dataset \cite{jia09}.

For each distorted image, we compute the outputs of masking-aware versions E-VGG, E-LPIPS, E-DISTS, and E-DeepWSD as proposed in \cite{cougalan2024enhancing}, due to their strong performance (see \refTbl{metricsResults}). We aggregate their scores using equal weights to avoid overfitting due to limited human-labeled data. We refer to this combined metric as \textit{Ensemble}, which leverages the complementary strengths of its components and achieves higher predictive accuracy than any single metric alone. This aligns with findings from \citet{lousal24}, where combining diverse IQA metrics led to more stable and reliable predictions.

%For each distorted image, we compute the outputs of masking-aware versions \cite{cougalan2024enhancing} of VGG, LPIPS, DISTS, and DeepWSD that are best performing in MOS prediction. 
%Each metric captures complementary aspects of quality degradation. 
%The Ensemble prediction combines the scores using equal weights because trained weight distributions overfitted to the data, given the limited size of available human labels.
%
%This ensemble-based strategy draws on the findings of \citet{lousal24}, showing that aggregating diverse IQA metrics results in more stable predictions. Interestingly, we observe that the combination of the metrics in an ensemble emphasizes the strengths of each component metric, leading to superior predictive accuracy compared to any single metric alone.
%
The resulting pseudo-MOS scores are used to train our multiscale quality metric and are critical to reaching state-of-the-art performance without requiring large-scale human annotations.

\subsection{Multiscale CNN Model}\label{sec:multiCNN}

We define our perceptual quality model as a multiscale convolutional neural network $F$ that estimates a pixel-wise perceptual mask $M \in [0,1]^{H \times W}$ and a global quality score $S_\mathrm{raw} \in \mathbb{R}$ for a given distorted image $Y \in \mathbb{R}^{H \times W \times 3}$ with respect to a reference image $X \in \mathbb{R}^{H \times W \times 3}$. 
The overall function of the network can be expressed as $(_\mathrm{raw}, M) = F(X, Y)$.
Unlike many recent deep FR-IQA approaches that rely on large pretrained backbones or increase network capacity, we follow a multiscale architecture to improve performance and interpretability without adding computational burden \cite{nah2017deep, ranjan17}. This design is inspired by prior work, which demonstrated that perceptual error prediction can be improved using spatial weighting informed by human visual masking \cite{cougalan2024enhancing, mantiuk2021fovvideovdp, mantiuk2011hdr}. We found a multiscale architecture to be a good choice for this task, as the different scales correspond to different spatial frequency bands. High-frequency distortions, such as impulse noise, may be better captured at finer scales, while low-frequency distortions, such as color changes, correspond more to coarser scales. 
% The multiscale network learns to identify these properties and adjusts the predicted visibility in accordance with the local spatial frequency of the image content.
However, in comparison with classical HVS models that are based on narrowband masking effects measured in controlled settings, natural images can exhibit complex correlations across scales. A multiscale architecture benefits from propagating coarse-level priors to prevent inconsistent visibility estimates between scales.

We apply a multi-level image pyramid using bicubic downscaling to both reference and distorted images:
\begin{equation}
\{X^l, Y^l\}_{l=1}^{L}, \quad \text{where } X^l, Y^l \in \mathbb{R}^{\frac{H}{2^{l-1}} \times \frac{W}{2^{l-1}} \times 3}
\end{equation}
Here, $l=1$ denotes the coarsest scale and $l=L$ the original resolution. Each scale level is processed by a shared convolutional subnetwork $\phi_l$ that finds a residual error mask $M_r$:
\begin{equation}
M_{r}^l = \phi_l(X^l, Y^l, M^{l-1})
\end{equation}
with $M^0 = 0$. 
To obtain the input mask for the next scale, the error mask of $l-1$ is combined with the output of  $\phi_l$ and upsampled using bilinear interpolation (Up):
\begin{equation}
M^l = \text{Up}\left( \phi_l(X^l, Y^l, M^{l-1}) + M^{l-1} \right)
\end{equation}
The residual mask is fed to all scales rather than computed independently per scale because it is not frequency-selective in the classical sense. Instead, it encodes spatially localized, content-dependent visibility as a joint function of local contrast, structure, and texture. Feeding this contextual information forward allows finer scales to leverage the coarse-scale visibility prediction without relearning it.

The CNN at each scale consists of five convolutional layers with channel sizes 16–32–64–32–16. All layers use ReLU activations, except the final layer, which uses a sigmoid to constrain the mask values in $[0,1]$. 
The mask of the finest resolution $M^L$ is used to modulate the local error of the input images:
\begin{equation}
\tilde{E} = M^L \odot E, \quad \text{where }  E = |X - Y| 
\label{eq:ML}
\end{equation}
where $\odot$ denotes element-wise multiplication. The final global error $S_{\mathrm{raw}}$ is aggregated over the finest scale:
\begin{equation}
S_{\mathrm{raw}} = \frac{1}{H W} \sum_{i,j} \tilde{E}_{i,j}
\end{equation}

To ensure comparability with the value range of MOS labels, $S_{\mathrm{raw}}$ and $\tilde{E}$ are normalized by a lightweight scalar mapper $G: \mathbb{R} \rightarrow \mathbb{R}$, similar to previous work \cite{zhang2018perceptual, cougalan2024enhancing}. The aggregated error $S_{\mathrm{raw}}$ is transformed into the predicted MOS score $MOS_\mathrm{predict}$, while the final visibility map is obtained by $M_V = G(\tilde{E})$. The small network $G$ is trained jointly with $F$ and learns a mapping from metric scale to perceptual scale.
% The overall prediction can therefore formally be formulated as:
% \begin{equation}
% F(X, Y) = \left(G\left(\frac{1}{HW} \sum_{i,j} M_{i,j} \cdot |X_{i,j} - Y_{i,j}|\right),\; M_v\right)
% \end{equation}

Compared with the single-scale version of this model (finest scale), incorporating multiple resolutions leads to significantly better results, especially when trained on augmented data. 
MILO is trained exclusively on pseudo-MOS scores generated by \textit{Ensemble} to guide the learning. Therefore, the training inputs are only the reference and distorted image and the pseudo-MOS score.
% We use the augmented training data generated via our pseudo-MOS strategy (see \refSec{mosPredict}), since the available human-labeled data is too sparse to let our model fully understand the relationship between distortion visibility and MOS score. 
For further training details, please refer to the supplementary material.

\subsection{Latent Space Masking}\label{sec:latentMasking}

Aside from RGB inputs, our perceptual masks can equally be derived from images decoded into latent space. Latents are essential for modern generative models like diffusion, which operate on latent representations for efficiency and guidance. Stable diffusion models \cite{Rombach_2022_CVPR}, for instance, perform generation in a compact latent space encoded via a variational autoencoder (VAE).
Direct latent-space operation reduces memory and compute overhead by avoiding repeated VAE decoding. Our perceptual masking further enhances this process by preserving spatial structure, enabling targeted restoration in semantically relevant regions. Processing the latent images requires no further modification of the pipeline.

\subsection{Curriculum Learning with Perceptual Masking}\label{sec:curriculum}
Image restoration models are typically trained using global quality scores or uniform loss functions, even though different regions of an image vary significantly in perceptual importance. Our perceptual mask addresses this limitation by providing spatially resolved information about distortion visibility. This enables a curriculum learning strategy \cite{bengio2009curriculum} in which the network first learns to restore less perceptually critical areas before focusing on regions where distortions are more visible. This gradual increase in task difficulty aligns better with how learning systems adapt and helps the network prioritize regions that matter most for perceived image quality.

We define curriculum learning over the course of training by introducing a scheduling parameter $\alpha \in [0, 1]$ that gradually shifts the loss emphasis from masked (easier, lower distortion visibility) regions to unmasked (difficult, perceptually salient) regions. As before, we define the reference and predicted images as $X$ and $Y$, respectively, and the perceptual mask by $M \equiv M^L$, which is normalized to $[0,1]$. The pixel-wise absolute difference is defined by 
$D = |X - Y|$.
Using $\alpha$, we define a curriculum-weighted L1 loss:
\begin{equation}
\mathcal{L}_{\text{curr}} = \left[ (1 - \alpha)(1 - M) \cdot D + \alpha M \cdot D \right]_{\text{mean}}
\end{equation}

%This loss initially prioritizes regions with low predicted visibility (easy content) and progressively emphasizes regions with high visibility (perceptual importance) as training advances. 
\noindent
The parameter $\alpha$ is scheduled as a function of the current training epoch $e$ and the total number of epochs $N$. We evaluate two scheduling strategies. The first one is a linear scheduling strategy: % defined as:
\begin{equation}
\alpha = \frac{e}{N}
\end{equation}
The second strategy modulates the emphasis by a cosine function: 
\begin{equation}
\label{eq:cl-cosine}
\alpha = \frac{1}{2} \left(1 - \cos\left(\pi \cdot \frac{e}{N} \right)\right)
\end{equation}
%
% Figure~\ref{fig:alpha_schedule} illustrates the progression of both schedules. 
While the linear schedule increases uniformly, the cosine variant introduces a slow start and accelerates mid-training, providing a smoother transition between easy and hard regions. With this flexibility, we aim to enable fine control over the difficulty progression, which we show to be beneficial for stable and perceptually aligned restoration performance in \refSec{lossfunction}.

% \karol{add perhaps some short remark, how in practice the reference image requirement can be avoided at the test time. $D$ requires the reference image. Or, it is obvious? Not obvious to me.}

%%%%%%%%%%%%%%%%%%%%%%%%%%%%%%%%%%%%%%%%%%%%%%%%%%%%%%%%%%%%%%%%%%%%%%%%%%

\mycfigure{mask_vis}{Visualization of residual masks and visibility maps in both image and latent space for two distortion types: Gaussian noise (top row) and blur (bottom row). In the Image Masks ($M^L$, \refEq{ML}), darker regions indicate stronger visual masking. For Gaussian noise, masking increases with contrast, resulting in reduced noise visibility in the high-contrast motorcycle region. Conversely, the bright motorcycle region in the blur example reflects weak masking, as blur is more noticeable along high-contrast edges. The Image Visibility Maps ($M_V$, \refFig{pipeline5}) capture the resulting visibility of these distortions. The Latent Masks and Latent Visibility Maps exhibit spatial and magnitude correlations with their image-space counterparts, though their patterns are less visually interpretable. 
Their smoother, more blurred appearance results from the compact image representation produced by the VAE encoder of  \citet{lin2024diffbir}, which required 8$\times$ upsampling for visualization in this figure.
}

\definecolor{lightgray}{RGB}{240, 240, 240}
\definecolor{darkgray}{RGB}{205, 205, 205}

\begin{table}[t]
    \caption{Comparison of our MILO metric against existing FR-IQA methods on three standard datasets. MILO$_\mathrm{I}$ and MILO$_\mathrm{L}$ denote the image-based and latent-based versions of our method, respectively. Numbers in parentheses indicate the number of reference images used for training (\# training pairs = \# reference images * 5 levels * 25 distortion types); * denotes training with only one random distortion at five distortion levels per reference (\# training pairs = \# reference images * 5 levels)---please see \refSec{ablation} for more details. Higher values indicate better quality prediction.}
    \vspace{-0.3cm}
    \label{tab:metricsResults}
    \setlength{\tabcolsep}{2pt}
    \renewcommand{\arraystretch}{1.15}
    \resizebox{\columnwidth}{!}{%
    \begin{tabular}{lccccccccccccc}
        \toprule
        & \multicolumn{3}{c}{\textbf{CSIQ}} & \multicolumn{3}{c}{\textbf{TID}}  & \multicolumn{3}{c}{\textbf{PIPAL}} \\
        \cmidrule(lr){2-4} \cmidrule(lr){5-7} \cmidrule(lr){8-10}
        \textbf{Metric} & PLCC & SRCC & KRCC & PLCC & SRCC & KRCC & PLCC & SRCC & KRCC \\

        \midrule
        % MAE rows
        % MAE & \cellcolor{lightgray} 0.819 & \cellcolor{lightgray} 0.800 & \cellcolor{lightgray} 0.599 & \cellcolor{lightgray} 0.627 & \cellcolor{lightgray} 0.545 & \cellcolor{lightgray} 0.409 & \cellcolor{lightgray} 0.401 & \cellcolor{lightgray} 0.450 & \cellcolor{lightgray} 0.312 & \cellcolor{lightgray} 0.458 & \cellcolor{lightgray} 0.443 & \cellcolor{lightgray} 0.304 \\

        PSNR & \cellcolor{lightgray} 0.851 & \cellcolor{lightgray} 0.837 \cellcolor{lightgray} & \cellcolor{lightgray} 0.645 & \cellcolor{lightgray} 0.726 & \cellcolor{lightgray} 0.714 & \cellcolor{lightgray} 0.540  &  \cellcolor{lightgray} 0.468 & \cellcolor{lightgray} 0.456 & \cellcolor{lightgray} 0.314 \\

        SSIM & \cellcolor{lightgray} 0.848 & \cellcolor{lightgray} 0.863 &\cellcolor{lightgray} 0.665 &\cellcolor{lightgray} 0.697 &\cellcolor{lightgray} 0.663 &\cellcolor{lightgray} 0.479 &         \cellcolor{lightgray} 0.550 & \cellcolor{lightgray} 0.534 &\cellcolor{lightgray} 0.373 \\

        FSIM & \cellcolor{lightgray} 0.900 & \cellcolor{lightgray} 0.913 & \cellcolor{lightgray} 0.740 & \cellcolor{lightgray} 0.847 & \cellcolor{lightgray} 0.789 & \cellcolor{lightgray} 0.611  &          \cellcolor{lightgray} 0.651 & \cellcolor{lightgray} 0.617 & \cellcolor{lightgray} 0.441 \\

        HDR-VDP-2 & \cellcolor{lightgray} 0.761 & \cellcolor{lightgray} 0.886 & \cellcolor{lightgray} 0.704 & \cellcolor{lightgray} 0.715 & \cellcolor{lightgray} 0.753 & \cellcolor{lightgray} 0.571 &  \cellcolor{lightgray} 0.514 &         \cellcolor{lightgray} 0.503 & \cellcolor{lightgray} 0.354 \\

        PieAPP & \cellcolor{lightgray} 0.827 & \cellcolor{lightgray} 0.840 & \cellcolor{lightgray} 0.653 & \cellcolor{lightgray} 0.832 & \cellcolor{lightgray} 0.849 & \cellcolor{lightgray} 0.652 &  \cellcolor{lightgray} 0.729 & \cellcolor{lightgray} 0.709 & \cellcolor{lightgray} 0.521 \\

        VGG & \cellcolor{lightgray} 0.938 & \cellcolor{lightgray} 0.952 & \cellcolor{lightgray} 0.804 & \cellcolor{lightgray} 0.853 & \cellcolor{lightgray} 0.820 & \cellcolor{lightgray} 0.639 &          \cellcolor{lightgray} 0.643 & \cellcolor{lightgray} 0.610 & \cellcolor{lightgray} 0.432 \\

        LPIPS & \cellcolor{lightgray} 0.944 & \cellcolor{lightgray} 0.929 & \cellcolor{lightgray} 0.769 & \cellcolor{lightgray} 0.803 & \cellcolor{lightgray} 0.756 & \cellcolor{lightgray} 0.568  &         \cellcolor{lightgray} 0.640 & \cellcolor{lightgray} 0.598 & \cellcolor{lightgray} 0.424 \\

        DISTS & \cellcolor{lightgray} 0.947 & \cellcolor{lightgray} 0.947 & \cellcolor{lightgray} 0.796 & \cellcolor{lightgray} 0.839 & \cellcolor{lightgray} 0.811 & \cellcolor{lightgray} 0.619 &  \cellcolor{lightgray} 0.645 & \cellcolor{lightgray} 0.626 & \cellcolor{lightgray} 0.445 \\

        DeepWSD & \cellcolor{lightgray} 0.949 & \cellcolor{lightgray} 0.961 & \cellcolor{lightgray} 0.821 & \cellcolor{lightgray} 0.879 & \cellcolor{lightgray} 0.861 & \cellcolor{lightgray} 0.674 &  \cellcolor{lightgray} 0.593 & \cellcolor{lightgray} 0.584 & \cellcolor{lightgray} 0.409 \\

        FovVideoVDP & \cellcolor{lightgray} 0.795  & \cellcolor{lightgray} 0.821 & \cellcolor{lightgray} 0.632 & \cellcolor{lightgray} 0.742 & \cellcolor{lightgray} 0.727 & \cellcolor{lightgray} 0.544 &  \cellcolor{lightgray} 0.565 & \cellcolor{lightgray} 0.509 & \cellcolor{lightgray} 0.358 \\

        ColorVideoVDP & \cellcolor{lightgray} 0.885 & \cellcolor{lightgray} 0.895 & \cellcolor{lightgray} 0.728 & \cellcolor{lightgray} 0.864 & \cellcolor{lightgray} 0.853 & \cellcolor{lightgray} 0.672 & \cellcolor{lightgray} 0.625 & \cellcolor{lightgray} 0.587 & \cellcolor{lightgray} 0.418  \\

        FLIP & \cellcolor{lightgray} 0.731 & \cellcolor{lightgray} 0.724 & \cellcolor{lightgray} 0.527 & \cellcolor{lightgray} 0.591 & \cellcolor{lightgray} 0.537 & \cellcolor{lightgray} 0.413 &  \cellcolor{lightgray} 0.498 &  \cellcolor{lightgray} 0.442 & \cellcolor{lightgray} 0.306 \\

        WaDIQaM & \cellcolor{lightgray} 0.909 & \cellcolor{lightgray} 0.923 & \cellcolor{lightgray} 0.827 & \cellcolor{lightgray} 0.857 & \cellcolor{lightgray} 0.854 & \cellcolor{lightgray} 0.655 & \cellcolor{lightgray} 0.665 &  \cellcolor{lightgray} 0.675 & \cellcolor{lightgray} 0.484 \\

        TOPIQ & \cellcolor{lightgray} 0.954 & \cellcolor{lightgray} 0.961 & \cellcolor{lightgray} 0.827 & \cellcolor{lightgray} 0.906 & \cellcolor{lightgray} 0.908 & \cellcolor{lightgray} 0.728 & \cellcolor{lightgray} 0.710 &  \cellcolor{lightgray} 0.693 & \cellcolor{lightgray} 0.503 \\

        \midrule
        % DISTS rows

        E-VGG & \cellcolor{lightgray} 0.943 & \cellcolor{lightgray} 0.942 & \cellcolor{lightgray} 0.784 & \cellcolor{lightgray} 0.909 & \cellcolor{lightgray} 0.892 & \cellcolor{lightgray} 0.714 &  \cellcolor{lightgray} 0.676 & \cellcolor{lightgray} 0.650 & \cellcolor{lightgray} 0.464 \\

        E-LPIPS  & \cellcolor{lightgray} 0.945 & \cellcolor{lightgray} 0.940 & \cellcolor{lightgray} 0.783 & \cellcolor{lightgray} 0.900 & \cellcolor{lightgray} 0.886 & \cellcolor{lightgray} 0.703 &  \cellcolor{lightgray} 0.707 & \cellcolor{lightgray} 0.675 & \cellcolor{lightgray} 0.489 \\
        
        E-DISTS  & \cellcolor{lightgray} 0.930 & \cellcolor{lightgray} 0.916 & \cellcolor{lightgray} 0.743 & \cellcolor{lightgray} 0.910 & \cellcolor{lightgray} 0.899 & \cellcolor{lightgray} 0.717 &  \cellcolor{lightgray} 0.729 & \cellcolor{lightgray} 0.701 & \cellcolor{lightgray} 0.509 \\

        E-DeepWSD  & \cellcolor{lightgray} 0.932 & \cellcolor{lightgray} 0.934 & \cellcolor{lightgray} 0.769 & \cellcolor{lightgray} 0.903 & \cellcolor{lightgray} 0.892 & \cellcolor{lightgray} 0.710 &  \cellcolor{lightgray} 0.665 & \cellcolor{lightgray} 0.632 & \cellcolor{lightgray} 0.450 \\

        Ensemble & \cellcolor{lightgray} 0.955 & \cellcolor{lightgray} 0.952 & \cellcolor{lightgray} 0.805 & \cellcolor{darkgray} \textbf{0.921} & \cellcolor{darkgray} \textbf{0.911} & \cellcolor{darkgray} \textbf{0.738} &  \cellcolor{lightgray} 0.734 & \cellcolor{lightgray} 0.704 & \cellcolor{lightgray} 0.513 \\

        \midrule
    
        % MILO$_\mathrm{I}$  & \cellcolor{lightgray} 0.891 & \cellcolor{lightgray} 0.898 & \cellcolor{lightgray} 0.707 & \cellcolor{lightgray} 0.867 & \cellcolor{lightgray} 0.856 & \cellcolor{lightgray} 0.665 & \cellcolor{lightgray} 0.648 & \cellcolor{lightgray} 0.636 & \cellcolor{lightgray} 0.452 \\

        % MILO$_\mathrm{I}$ (kadid) & \cellcolor{lightgray} 0.950 & \cellcolor{lightgray} 0.954 & \cellcolor{lightgray} 0.807 & \cellcolor{lightgray} 0.862 & \cellcolor{lightgray} 0.848 & \cellcolor{lightgray} 0.654 &  \cellcolor{lightgray} 0.710 & \cellcolor{lightgray} 0.677 & \cellcolor{lightgray} 0.490 \\

        MILO$_\mathrm{I}$ (1k) & \cellcolor{lightgray} 0.954 & \cellcolor{lightgray} 0.961 & \cellcolor{lightgray} 0.823 & \cellcolor{lightgray} 0.887 & \cellcolor{lightgray} 0.869 & \cellcolor{lightgray} 0.680 &  \cellcolor{lightgray} 0.730 & \cellcolor{darkgray} \textbf{0.711} & \cellcolor{darkgray} \textbf{0.519} \\

        % MILO$_\mathrm{I}$$_{multiscale-3k}$ & \cellcolor{lightgray} \textbf{0.961} & \cellcolor{lightgray} \textbf{0.965} & \cellcolor{lightgray} \textbf{0.830} & \cellcolor{lightgray} 0.882 & \cellcolor{lightgray} 0.864 & \cellcolor{lightgray} 0.674 & \cellcolor{lightgray} \textbf{0.949} & \cellcolor{lightgray} \textbf{0.948} & \cellcolor{lightgray} \textbf{0.801} & \cellcolor{lightgray} \textbf{0.747} & \cellcolor{lightgray} \textbf{0.719} & \cellcolor{lightgray} \textbf{0.528} \\

        MILO$_\mathrm{I}$ (10k) & \cellcolor{lightgray} 0.965 & \cellcolor{lightgray} 0.967 & \cellcolor{lightgray} 0.834 & \cellcolor{lightgray} 0.885 & \cellcolor{lightgray} 0.866 & \cellcolor{lightgray} 0.677 &  \cellcolor{lightgray} 0.731 & \cellcolor{lightgray} 0.708 & \cellcolor{lightgray} 0.516 \\

        MILO$_\mathrm{I}$ (20k)  & \cellcolor{darkgray} \textbf{0.967} & \cellcolor{darkgray} \textbf{0.968} & \cellcolor{darkgray} \textbf{0.838} & \cellcolor{lightgray} 0.885 & \cellcolor{lightgray} 0.866 & \cellcolor{lightgray} 0.678 &  \cellcolor{lightgray} 0.731 & \cellcolor{lightgray} 0.705 & \cellcolor{lightgray} 0.513 \\

        MILO$_\mathrm{I}$ (50k)  & \cellcolor{lightgray} 0.966 & \cellcolor{lightgray} 0.967 & \cellcolor{lightgray} 0.835 & \cellcolor{lightgray} 0.888 & \cellcolor{lightgray} 0.871 & \cellcolor{lightgray} 0.683 &  \cellcolor{darkgray} \textbf{0.736} & \cellcolor{darkgray} \textbf{0.711} & \cellcolor{darkgray} \textbf{0.519} \\

        MILO$_\mathrm{I}$ (100k*) & \cellcolor{lightgray} 0.962 & \cellcolor{lightgray} 0.962 & \cellcolor{lightgray} 0.824 & \cellcolor{lightgray} 0.885 & \cellcolor{lightgray} 0.860 & \cellcolor{lightgray} 0.673 &  \cellcolor{lightgray} 0.728 & \cellcolor{lightgray} 0.701 & \cellcolor{lightgray} 0.509 \\

        MILO$_\mathrm{I}$ (1M*)  & \cellcolor{lightgray} 0.961 & \cellcolor{lightgray} 0.959 & \cellcolor{lightgray} 0.818 & \cellcolor{lightgray} 0.897 & \cellcolor{lightgray} 0.876 & \cellcolor{lightgray} 0.692 &  \cellcolor{lightgray} 0.713 & \cellcolor{lightgray} 0.690 & \cellcolor{lightgray} 0.499 \\

        \midrule

        % MILO$_\mathrm{L}$ (kadid) & \cellcolor{lightgray} 0.909 & \cellcolor{lightgray} 0.918 & \cellcolor{lightgray} 0.742 & \cellcolor{lightgray} 0.827 & \cellcolor{lightgray} 0.814 & \cellcolor{lightgray} 0.613 &  \cellcolor{lightgray} 0.552 & \cellcolor{lightgray} 0.532 & \cellcolor{lightgray} 0.370 \\

        MILO$_\mathrm{L}$ (100k*)   & \cellcolor{lightgray} 0.963 & \cellcolor{lightgray} 0.958 & \cellcolor{lightgray} 0.817 & \cellcolor{lightgray} 0.877 & \cellcolor{lightgray} 0.856 & \cellcolor{lightgray} 0.668 &  \cellcolor{lightgray} 0.714 & \cellcolor{lightgray} 0.678 & \cellcolor{lightgray} 0.489 \\

        MILO$_\mathrm{L}$ (500k*) & \cellcolor{lightgray} 0.960 & \cellcolor{lightgray} 0.957 & \cellcolor{lightgray} 0.814 & \cellcolor{lightgray} 0.878 & \cellcolor{lightgray} 0.860 & \cellcolor{lightgray} 0.671 &  \cellcolor{lightgray} 0.715 & \cellcolor{lightgray} 0.684 & \cellcolor{lightgray} 0.495 \\

        MILO$_\mathrm{L}$ (1M*)  & \cellcolor{lightgray} 0.963 & \cellcolor{lightgray} 0.960 & \cellcolor{lightgray} 0.821 & \cellcolor{lightgray} 0.881 & \cellcolor{lightgray} 0.862 & \cellcolor{lightgray} 0.675 & \cellcolor{lightgray} 0.711 & \cellcolor{lightgray} 0.683 & \cellcolor{lightgray} 0.493 \\
        
        \bottomrule

    \end{tabular}}
\end{table}

\section{Results}

To validate the performance of our metric, we present quantitative (Sec.~\ref{sec:evaluation}--\ref{sec:speed}) and qualitative results (Sec.~\ref{sec:qualityMask}).

\subsection{Predictions of Image Quality}\label{sec:evaluation}
Table \ref{tab:metricsResults} presents the correlations between predicted MOS values and human-labeled ground truth scores across standard FR-IQA datasets. We compare our metric against both traditional and learning-based approaches. Please see the supplementary for more information about the experimental setup. 
% The augmented training data (amount of reference images given in parentheses) of MILO does not include any images of the original KADID dataset.
%
Traditional metrics such as MAE, SSIM, HDR-VDP-2, and ColorVideoVDP show consistently lower performance in predicting perceptual image quality. These methods, while computationally efficient, rely on handcrafted assumptions and may not account for spatial context or complex perceptual phenomena. In contrast, our approach delivers substantially higher correlation with human ratings across all benchmarks.
Compared to learning-based metrics, like LPIPS or DeepWSD, our method achieves superior prediction accuracy. Despite relying on a significantly smaller network, MILO outperforms other methods by combining targeted architecture design with semantically meaningful data and an effective augmentation strategy (see \refSec{mosPredict}).

We additionally evaluate our proposed \textit{Ensemble}, a predictor based on refined error metrics from recent work \cite{cougalan2024enhancing}. \textit{Ensemble} was used as the basis of our pseudo-MOS generation. The results demonstrate how \textit{Ensemble} can achieve strong performance by combining the strengths of its components. Remarkably, although MILO is trained on the \textit{Ensemble} outputs, it generalizes better to unseen data, such as the extensive PIPAL benchmark, where most distortions arise from undertrained neural networks that produce non-deterministic and unpredictable artifacts. This indicates that MILO's multiscale network can capture perceptual structure beyond the representational scope of its training labels.

% On the KADID test set, both our metric and the ensemble predictor are on par and achieve almost perfect scores. However, while doing so, our metric is significantly faster at inference time given its lightweight architecture. These results, discussed further in the following paragraph, make our model well-suited for time-critical applications and use as a perceptual loss (see \refSec{lossfunction}).

\subsection{Runtime Performance}\label{sec:speed}

We further compare the learning-based metrics' inference time, averaging 1000 runs (NVIDIA Quadro RTX 8000 GPU). Images have a resolution of 512$\times$384, and we consider the pure metric processing time (no I/O).
As shown in \refTbl{avg_inference_ms}, our method achieves significantly faster inference speed, with an average processing time of only 3.94\,ms per image. 
%This makes MILO over 18$\times$ faster than DeepWSD and more than 90$\times$ faster than TOPIQ, highlighting its ideal use in real-time or large-scale applications. 
Notably, this remarkable speed advantage is achieved without compromising accuracy (see \refSec{evaluation}). We also observe that the latent version of our metric (MILO$_\mathrm{L}$ ) is faster than its image-space counterpart (MILO$_\mathrm{I}$), due to the reduced dimensionality of the compressed latent representation.

\begin{table}[t]
\centering
\footnotesize
\caption{Average inference time per image for learning-based FR metrics.}
\setlength{\tabcolsep}{7pt}
\vspace{-0.3cm}
\begin{tabular}{l p{1.16cm} p{1.16cm} p{1.16cm} p{1.16cm}}
\toprule
\textbf{Method} & PieApp & DeepWSD & LPIPS & DISTS \\
\midrule
\textbf{Time (ms)} & \cellcolor{lightgray} 72.75 & \cellcolor{lightgray} 24.53 & \cellcolor{lightgray} 18.97 & \cellcolor{lightgray} 21.03 \\
\bottomrule
\end{tabular}
\begin{tabular}{llllll}
\toprule
\textbf{Method} & TOPIQ & WaDIQaM & Ensemble & MILO$_\mathrm{I}$ & MILO$_\mathrm{L}$ \\
\midrule
\textbf{Time (ms)} & \cellcolor{lightgray} 357.69 & \cellcolor{lightgray} 172.06 & \cellcolor{lightgray} 105.40 & \cellcolor{darkgray} \textbf{3.94} & \cellcolor{darkgray} \textbf{2.16} \\
\bottomrule
\end{tabular}
\label{tab:avg_inference_ms}
\end{table}

\subsection{Quality of the Visibility Map}\label{sec:qualityMask}
\refFig{mask_vis} compares the visibility maps and residual masks produced by MILO$_\mathrm{I}$ and MILO$_\mathrm{L}$ to evaluate the interpretability and perceptual relevance of the learned error weighting. 
%We consider the same image with different distortions. 
%
% The maps are obtained by applying the learned spatial weighting to the local mean absolute error, followed by their transformation into the MOS domain using the trained scalar network $G$. This procedure ensures that the output reflects perceptual distortion intensity in units that are directly comparable to human judgments.
%
% Since the raw outputs of many traditional and learning-based IQA metrics vary in scale and dynamic range, a direct visual comparison would be misleading. Therefore, we apply the same scalar mapping procedure to the error maps of baseline metrics, enabling a fair and perceptually grounded comparison by aligning their outputs equally with the MOS scale. This procedure was inspired by related work \cite{cougalan2024enhancing} and ensures that all visualized responses can be interpreted relative to perceptual quality degradation.
%
%In contrast to most previous metrics, our model is not constrained to modulate an existing score but learns to generate its own spatial weighting from scratch. The resulting visibility masks align well with known perceptual patterns. For instance, in the case of Gaussian noise (top row), the model suppresses uniformly distributed pixel-level error and highlights only perceptually salient disruptions. For blur (bottom row), the visibility map emphasizes edge degradation and loss of fine structure.
We consider the same image subjected to two different distortions: Gaussian noise (top row) and blur (bottom row). As expected, noise visibility is more strongly masked in the high-contrast motorcycle region than in the low-contrast sky. Conversely, blur is more perceptible along the high-contrast edges of the motorcycle pattern.

These results suggest that the network has implicitly learned the link between local error visibility and perceptual masking, without requiring explicit supervision. The learned mask acts as a visibility estimator that reflects both the type and spatial structure of the distortion in context with the image content. 
Notably, the network learns to generate per-pixel visibility maps using only per-image pseudo-MOS supervision. While a similar insight was reported in \cite{cougalan2024enhancing}, our approach replaces human MOS annotations with synthetic pseudo-MOS data. Leveraging large-scale training on this data, our lightweight multi-scale metric achieves significantly improved image quality prediction accuracy.

\section{Applications}\label{sec:lossfunction}
Beyond image quality prediction, our metric can be applied directly as a perceptual loss in applications that aim for image restoration or optimization. Unlike conventional loss functions (e.g., L1 or SSIM loss), which treat all image regions equally, our approach provides spatially varying guidance by the perceptual mask $M^L$. Through guidance, we allow restoration networks to prioritize learning in perceptually important regions.
This spatial weighting facilitates curriculum learning that adapts the training objective over time, guiding the network to first solve perceptually easier subproblems before addressing more challenging regions.
In the following, we demonstrate the practical applicability of our metric to improve on many image restoration and diffusion optimization tasks. For extensive visual results, please also see the provided HTML.

%%%%%%%%%%%%%%%%%%%%%%%%%%%%%%%%%%%%%%%%%%%%%%%%%%%%%%%%%%%%
\subsection{Loss for Image Restoration}
%%%%%%%%%%%%%%%%%%%%%%%%%%%%%%%%%%%%%%%%%%%%%%%%%%%%%%%%%%%%
We evaluate the applicability of our metric as a perceptual loss in image restoration tasks.  
To this end, we integrate it as the training objective in the common Restormer framework \cite{zamir2022restormer} and test for restoration performance in image denoising and motion deblurring.  
% All models are trained under identical conditions, including fixed learning rate and number of iterations, to ensure comparability.

% \begin{table*}[]
%     \setlength{\tabcolsep}{4pt}
%     \caption{Comparison of \texttt{MAE\_random\_full} and \texttt{EMAEMScale3} models for blind Gaussian denoising at different noise levels ($\sigma$). We report PSNR, SSIM, and LPIPS metrics.}
%     \resizebox{\textwidth}{!}{%
%     \begin{tabular}{lccc|ccc|ccc}
%         \toprule
%         \textbf{Method} & 
%         \multicolumn{3}{c|}{$\sigma = 15$} & 
%         \multicolumn{3}{c|}{$\sigma = 25$} & 
%         \multicolumn{3}{c}{$\sigma = 50$} \\
%         \cmidrule(lr){2-4} \cmidrule(lr){5-7} \cmidrule(lr){8-10}
%          & PSNR$\uparrow$ & SSIM$\uparrow$ & LPIPS$\downarrow$ 
%          & PSNR$\uparrow$ & SSIM$\uparrow$ & LPIPS$\downarrow$
%          & PSNR$\uparrow$ & SSIM$\uparrow$ & LPIPS$\downarrow$ \\
%         \midrule
%         \texttt{MAE\_random\_full} & 
%         \textbf{34.83} & \textbf{0.943} & 0.052 &
%         \textbf{32.36} & \textbf{0.910} & 0.087 &
%         \textbf{29.21} & \textbf{0.842} & 0.160 \\
        
%         \texttt{EMAEMScale3} & 
%         34.35 & 0.941 & \textbf{0.046} &
%         31.86 & 0.907 & \textbf{0.075} &
%         28.70 & 0.837 & \textbf{0.136} \\
%         \bottomrule
%     \end{tabular}}
% \label{tbl:denoising_comparison}
% \end{table*}

\subsubsection{Image Denoising}
We train on BSD400 \cite{MartinFTM01} with additive Gaussian noise ($\sigma \in [0, 50]$) and evaluate on five standard benchmarks from Restormer \cite{zamir2022restormer}. \refFig{app_denoising} shows that our method yields sharper, more detailed outputs than the baseline.
\refTbl{denoising_comp} compares seven training variants: $\ell_1$ loss baseline, two full-reference setups using all 25 distortions, two with increased reference diversity but only one distortion per image at five distortion levels, and two curriculum learning variants (1M* model with linear and cosine scheduling).
Our loss consistently improves perceptual quality.
Notably, using more reference images—even with fewer distortions per image—improves generalization. While this benefit was not reflected in the classic metric evaluation scores like PLCC or SRCC (\refTbl{metricsResults}), it becomes evident in the downstream restoration performance.
Curriculum learning provides further gains, with the cosine scheduling being in slight favour.

\subsubsection{Motion Deblurring}
For motion deblurring, we use the GoPro dataset \cite{nah2017deep} for both training and evaluation.  
We apply only the best-performing version of our metric from the previous experiments---trained with 1M different images at five distortion levels and cosine curriculum learning---as it showed clear advantages over other configurations.  
Visual comparisons are provided in \refFig{app_deblurring}.
Supervision with our metric leads to significantly better performance compared to $\ell_1$ loss results (see \refTbl{deblurring_results}).

%%%%%%%%%%%%%%%%%%%%%%%%%%%%%%%%%%%%%%%%%%%%
\subsection{Latent Space Optimization}
%%%%%%%%%%%%%%%%%%%%%%%%%%%%%%%%%%%%%%%%%%%%
%
\begin{table}[t]
    \caption{Denoising results at different noise levels ($\sigma$) measured by SSIM, LPIPS and our \textit{Ensemble} metrics. The baseline uses MAE loss. Numbers in parentheses show the number of training references; * indicates one random distortion at five distortion levels per reference (instead of the full set of 25 distortion types)---please see \refSec{ablation} for more details. Curriculum learning (curr) uses the \textit{MILO (1M*)} model.}
    \vspace{-0.3cm}
    \setlength{\tabcolsep}{4pt}
    \renewcommand{\arraystretch}{1.15}
    \resizebox{\columnwidth}{!}{%
    \begin{tabular}{lcccccc}
        \toprule
        \textbf{Method} & 
        \multicolumn{3}{c}{$\sigma = 25$} & 
        \multicolumn{3}{c}{$\sigma = 50$} \\
        \cmidrule(lr){2-4} \cmidrule(lr){5-7}
          & SSIM$\uparrow$ & LPIPS$\downarrow$ & Ens.$\downarrow$
          & SSIM$\uparrow$ & LPIPS$\downarrow$ & Ens.$\downarrow$ \\
        \midrule
        \text{MAE} & 
        \cellcolor{lightgray} 0.9046 & \cellcolor{lightgray} 0.0966 & \cellcolor{lightgray} 0.2582 &
        \cellcolor{lightgray} 0.8342 & \cellcolor{lightgray} 0.1686 & \cellcolor{lightgray} 0.3615 \\

        MILO (10k) & 
        \cellcolor{lightgray} 0.9038 & \cellcolor{lightgray}  0.0887 & \cellcolor{lightgray} 0.2472 &
        \cellcolor{lightgray} 0.8331 & \cellcolor{lightgray}  0.1525 & \cellcolor{lightgray} 0.3458 \\

        % MILO (20k) & 
        % 34.28 & 0.9385 & 0.0533 & 0.1935 &
        % 31.83 & 0.9039 & 0.0850 & 0.2474 &
        % 28.68 & 0.8323 & 0.1484 & 0.3480 \\

        MILO (50k) & 
        \cellcolor{darkgray} \textbf{0.9050} & \cellcolor{lightgray} 0.0837 & \cellcolor{lightgray} 0.2493 &
        \cellcolor{lightgray} 0.8337 & \cellcolor{lightgray} 0.1432 & \cellcolor{lightgray} 0.3487 \\

        MILO (100k*) & 
        \cellcolor{lightgray} 0.9040 & \cellcolor{lightgray} 0.0834 & \cellcolor{lightgray} 0.2451 &
        \cellcolor{lightgray} 0.8333 & \cellcolor{lightgray} 0.1451 & \cellcolor{lightgray} 0.3427 \\

        MILO (1M*) & 
        \cellcolor{lightgray} 0.9038 & \cellcolor{lightgray} 0.0766 & \cellcolor{lightgray} 0.2385 &
        \cellcolor{lightgray} 0.8332 & \cellcolor{lightgray} 0.1340 & \cellcolor{lightgray} 0.3373 \\

        MILO (curr$_{lin}$) & 
        \cellcolor{lightgray} 0.9037 & \cellcolor{lightgray} 0.0717 & \cellcolor{lightgray} 0.2325 &
        \cellcolor{lightgray} 0.8330 & \cellcolor{lightgray} 0.1282 & \cellcolor{lightgray} 0.3324 \\

        MILO (curr$_{cos}$) & 
        \cellcolor{lightgray} 0.9028 & \cellcolor{darkgray} \textbf{0.0691} & \cellcolor{darkgray} \textbf{0.2323} & 
        \cellcolor{darkgray} \textbf{0.8309} & \cellcolor{darkgray} \textbf{0.1256} & \cellcolor{darkgray} \textbf{0.3313} \\
        \bottomrule
    \end{tabular}}
\label{tab:denoising_comp}
\end{table}
Latent-space optimization has become a powerful strategy for guiding generative models, especially in the context of diffusion-based image restoration. DiffBIR \cite{lin2024diffbir} is a recent state-of-the-art blind image restoration framework. It decouples the restoration process into two stages. First, a task-specific restoration module removes degradations in the image. 
Second, a generative diffusion model, IRControlNet \cite{lin2024diffbir}, reconstructs perceptual details in the latent space using stable diffusion.

We extend DiffBIR by integrating MILO$_\mathrm{L}$ (1M* version) into the framework's second stage. During inference, MILO operates as a latent-space guidance mechanism for the optimization process. Unlike the original region-adaptive MSE guidance of DiffBIR, which requires decoding the latent representation to RGB space at each step, our method applies masking directly in latent space. This avoids repeated VAE decoding and enables faster, memory-efficient optimization. 
MILO computes a spatial distortion mask ($M^L$) based on perceptual differences between the latent representations of the distorted image and the cleaned image of the restoration module of the first stage. This mask is then used to modulate the loss function with a curriculum learning scheme (\refEq{cl-cosine}, \refSec{curriculum}), progressively focusing training on perceptually salient regions. 
Our method requires no retraining of the diffusion module and integrates seamlessly into the existing inference process of DiffBIR.
% Compared to the original region-adaptive MSE loss, which is limited to pixel space and computationally costly, MILO-L offers a principled and efficient alternative that aligns well with the semantics of the latent space used in diffusion-based generation.

We apply our approach to multiple blind restoration tasks: denoising, super-resolution, and face restoration. 
Visual comparisons are provided in \refFig{merged_diff} and in the supplementary material.
For quantitative analysis, we employ the no-reference metrics TOPIQ \cite{chen2024topiq}, CLIP-IQA \cite{wang2023exploring}, and MUSIQ \cite{ke2021musiq} since no reference image is available during inference. Results in \refTbl{apps_results} are listed as an average score for all datasets. For more detailed analysis of different noise levels and individual datasets, please refer to the supplementary material.

\begin{table}[t]
\centering
\caption{Quantitative results for the motion deblurring task. Our method employs a curriculum learning scheme and is compared to the baseline MAE loss.}
\vspace{-0.3cm}
\begin{tabular}{lcccc}
\toprule
\textbf{Model} & \textbf{PSNR} ↑ & \textbf{SSIM} ↑ & \textbf{LPIPS} ↓ & \textbf{Ensemble} ↓ \\
\midrule
MAE & \cellcolor{lightgray} 29.1441 & \cellcolor{lightgray} 0.8862 & \cellcolor{lightgray} 0.1436 & \cellcolor{lightgray} 0.5600 \\
MILO & \cellcolor{darkgray} \textbf{29.6523} & \cellcolor{darkgray} \textbf{0.8965} & \cellcolor{darkgray} \textbf{0.1293} & \cellcolor{darkgray} \textbf{0.5462} \\
\bottomrule
\end{tabular}
\label{tab:deblurring_results}
\end{table}

\myccfigure{app_denoising}{Visual comparison for the denoising task using Restormer. Our metric-based supervision preserves sharper details and reduces oversmoothing compared to training with standard $\ell_1$ loss. The denoising with Restormer gives more importance to the darker regions, and as a result, we can get sharper details under low-light conditions. Here, we apply local tonemapping \cite{Farbman08} to the cropped insets for better visualisation.}{0.88}
\vspace{-0.2cm}
\myccfigure{app_deblurring}{Visual comparison for the deblurring task using Restormer.}{0.88}

\begin{table*}[t]
    \setlength{\tabcolsep}{15pt}
    \caption{Quantitative results for the three diffusion-based blind optimization tasks evaluated by established no-reference metrics. The baseline is the standard DiffBIR implementation. Here, we use the 1M* model of MILO with curriculum learning.}
    \vspace{-0.3cm}
    \resizebox{\textwidth}{!}{%
    \begin{tabular}{lcccccccccccc}
        \toprule
        \textbf{Method} & 
        \multicolumn{3}{c}{Denoising} & 
        \multicolumn{3}{c}{Super-Resolution} & 
        \multicolumn{3}{c}{Face Restoration} \\
        \cmidrule(lr){2-4} \cmidrule(lr){5-7} \cmidrule(lr){8-10}
         & TOPIQ$\uparrow$ & CLIP-IQA$\uparrow$ & MUSIQ$\uparrow$  
         & TOPIQ$\uparrow$ & CLIP-IQA$\uparrow$ & MUSIQ$\uparrow$ 
         & TOPIQ$\uparrow$ & CLIP-IQA$\uparrow$ & MUSIQ$\uparrow$  \\
		 
        \midrule
		
        DiffBIR & 
        \cellcolor{lightgray} 0.6651 & \cellcolor{lightgray} 0.6895 & \cellcolor{lightgray} 68.3026 &
        \cellcolor{lightgray} 0.6685 & \cellcolor{lightgray} 0.7238 & \cellcolor{lightgray} 68.9008 &
        \cellcolor{darkgray} \textbf{0.7584} & \cellcolor{darkgray} \textbf{0.8015} & \cellcolor{darkgray} \textbf{75.8715} \\

        MILO & 
        \cellcolor{darkgray} \textbf{0.6731} & \cellcolor{darkgray} \textbf{0.7097} & \cellcolor{darkgray} \textbf{69.0815} &
        \cellcolor{darkgray} \textbf{0.6997} & \cellcolor{darkgray} \textbf{0.7529} & \cellcolor{darkgray} \textbf{70.1442} &
        \cellcolor{lightgray} 0.6933 & \cellcolor{lightgray} 0.7516 & \cellcolor{lightgray} 75.8026 \\

        \bottomrule
    \end{tabular}}
\label{tab:apps_results}
\end{table*}
\begin{figure*}[h]
    \centering
    \includegraphics*[height = .91\textheight]{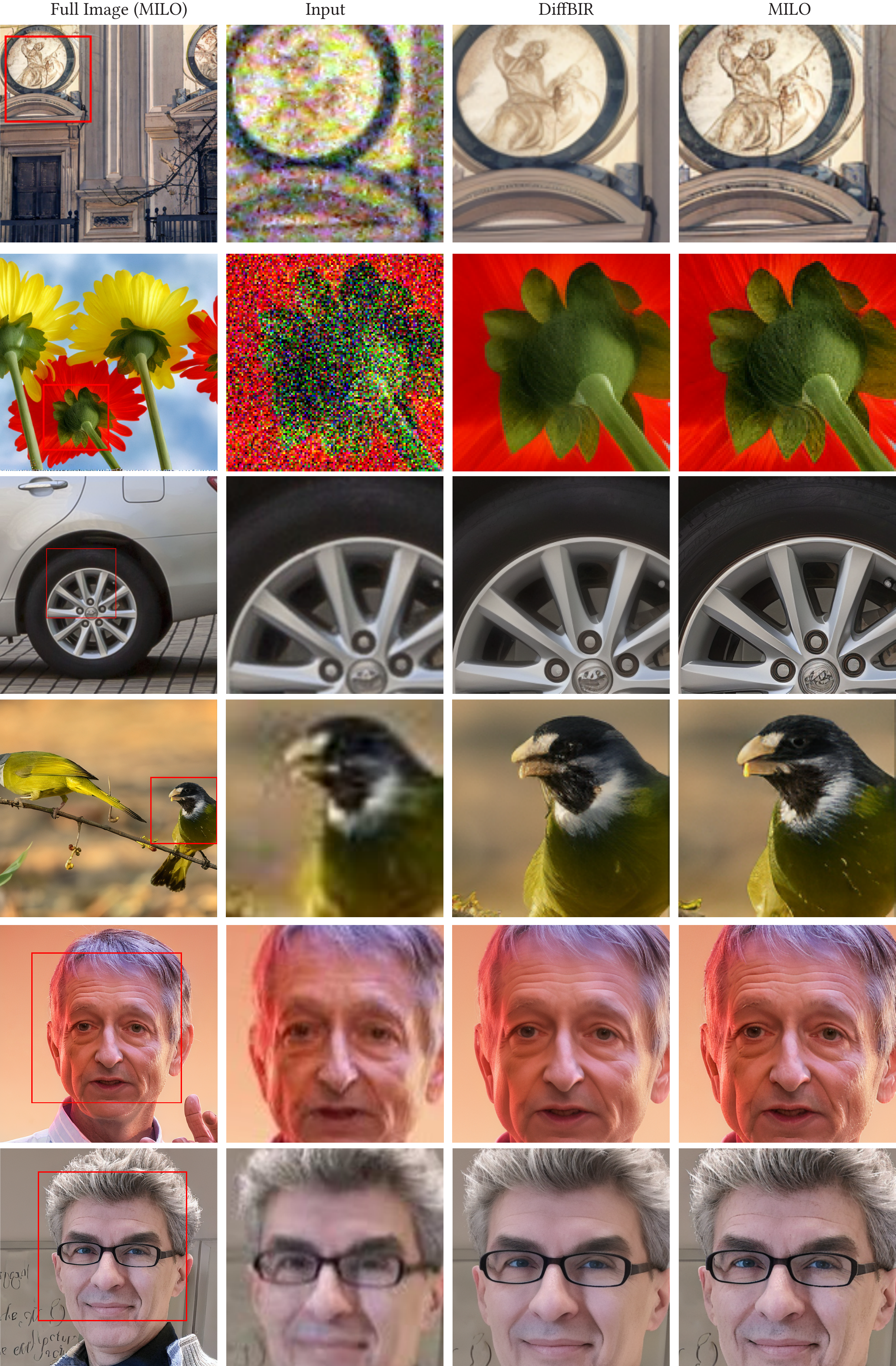}
    \caption{Visual comparison of using MILO for perceptually guided optimization in comparison to the original loss of DiffBIR \cite{lin2024diffbir}. Comparisons are given for diffusion-based blind denoising (first two rows), blind super-resolution (two middle rows), and face restoration (last two rows).}
    \label{fig:merged_diff}
\end{figure*}

\subsubsection{Blind Denoising}
For the blind diffusion-based image denoising, we evaluate our method on five standard datasets: Set12 \cite{zhang2017beyond}, BSD68 \cite{martin2001database}, Urban100 \cite{huang2015single}, Kodak24 \cite{franzen1999kodak}, and McMaster \cite{zhang2011color}. Here, we show the results for additive white Gaussian noise with $\sigma = 50$. 
Visually, MILO preserves sharper structures and generates more detailed textures. In contrast, the outputs from the MSE-based guidance tend to appear oversmoothed and lose high-frequency detail.
Quantitative results further support these observations. Our method consistently outperforms the baseline in all metrics, supporting the qualitative observations of improved quality.

\subsubsection{Super-Resolution}
We evaluate blind super-resolution using the three real-world datasets from prior work: the real image dataset of DiffBIR \cite{lin2024diffbir}, the RealSR dataset \cite{cai19}, and the DIV2K-val dataset \cite{agustsson17}. 
The qualitative results show that our method recovers finer details and better contrast. Edges appear sharper and textures more defined, whereas the original MSE-based guidance produces slightly blurrier outputs, partially lacking local detail.
The quantitative evaluations confirm the visual observations. Our method outperforms the baseline across all datasets and enables more perceptually accurate restoration without retraining of the diffusion pipeline or runtime compromises.

\subsubsection{Face Restoration}
We evaluate face restoration performance using the LFW \cite{huang08} and Wider \cite{zhou2022towards} datasets.
Visually, a clear difference can be seen. Our method preserves facial structure more accurately, enhances fine-scale details such as wrinkles and facial hair, and avoids the over-smoothing observed with MSE-based loss.
Surprisingly, quantitative results in \refTbl{apps_results} % \refTbl{face_restoration_results} 
do not support these findings. Although we could not observe visually objectionable failure cases, the no-reference metrics get worse, which is a known problem in the domain~\cite{voznesensky2022no, hu2025iqpfr}.

%%%%%%%%%%%%%%%%%%%%%%%%%%%%%%%%%%%%%%%%%%%%%%%%%%%%%
\section{Ablations}\label{sec:ablation}
%%%%%%%%%%%%%%%%%%%%%%%%%%%%%%%%%%%%%%%%%%%%%%%%%%%%%
In addition to ablations discussed before, such as the scheduling scheme of curriculum learning or the image domain of the metric (RGB vs. latent), we report the results of two further experiments.

\paragraph{Influence of Dataset Size.}
To evaluate the influence of dataset size, we generated synthetic training data from arbitrary images using our distortion forward model and ensemble labeling strategy. As shown in \refTbl{metricsResults}, increasing the number of reference images also improves the performance of the metric. For static metric correlation analysis, performance saturates after about 10,000 reference images. However, this saturation is not observed when MILO is used as a perceptual loss in downstream tasks such as denoising. Interestingly, in this setting, performance continues to improve with more reference images, indicating that data diversity might be meaningful beyond the saturation point of IQA correlations.

\paragraph{All Distortions vs. Randomized Samples.}
We further examine the necessity to apply all distortions to each reference image. We compare full distortion coverage (25 distortions per reference) against having one randomly sampled distortion per reference and more references instead (see \refTbl{denoising_comp}). Results indicate an increase in restoration performance in the randomized scheme, while the metric correlations are comparable (see \refTbl{metricsResults}).
These results suggest that the diversity of reference images is more significant than full distortion training to achieve good restoration performance. Still, we found that it is important to apply all distortion levels for each distortion type, so that the network can learn non-linearity in distortion perception versus its magnitude (cf. Fig. 1 in the supplementary).

% \subsection{Data Augmentation with Diffusion}
% results of former idea here

\section{Discussion}

MILO presents a new, perceptually accurate image quality metric that leverages pseudo-MOS labeling, curriculum learning, and a multiscale CNN architecture.
In the following, we would like to discuss and give further intuition about the core behaviour and functionality of MILO.

\paragraph{Efficiency of the Pipeline}
The pipeline was shown to be effective despite its simplicity. We attribute this effectiveness to the fact that the CNN learns core behaviors of the HVS directly from data. For that, it does not require explicit psychophysical calibration. The multiscale residual masks effectively emulate spatial‐frequency dependencies in a multi-dimensional manner. In comparison with psychophysic-based metrics, MILO (like many other learnable metrics) is not conditioned by distance to display or display characteristics. In the wild, this information is often not available. Learning from diverse image statistics, therefore, makes the method more robust and adaptable.

Without sharing the mask between scales the network fails to converge, given the results of additional experiments. The guidance from the intermediate masks, is needed to follow the intended masking cues, and to make informed decisions during the processing stages.

Furthermore, MILO is designed to predict relative perceptual differences rather than absolute detection thresholds, which are more beneficial for restoration guidance.

\paragraph{Relation of the Design to the Human Visual System}
The overall pipeline design is inspired by and aligns with established properties of the human visual system (HVS), though it remains simpler than full psychophysical models. The pyramid structure follows the HVS principle of coarse‑to‑fine analysis. 
MILO further reflects spatially selective integration by modulating local error before pooling. This follows human behavior, not integrating error uniformly, but instead being attracted to perceptually salient regions. When collecting MOS data, which we replicate in our pseudo-MOS data, strong distortions are likely to attract attention, modulated by their perceptibility due to visual masking. Near-threshold distortions are more likely to be perceived in salient image content that is more likely to be actively attended. Unlike fixed hand‑crafted models, where edges are always strong maskers, MILO’s learned masking is distortion‑dependent: edges mask noise effectively but do not mask blur, where distortions become even more visible (see \refFig{mask_vis}). This adaptive behavior better matches how the HVS responds under different distortion types.
\paragraph{Edge-enhanced reconstruction}
The reconstructed images of restoration, particularly from denoising and deblurring, show a behaviour similar to edge sharpening and higher-contrast.
We hypothesize that in our pseudo-MOS data, sharp edges and higher-contrast textures are promoted, leading to higher MOS values. This behavior could be learned from data involving blur distortion or compression artifacts, where sharper regions are associated with better-perceived quality. Note that MILO does not support naive contrast enhancement---for example, the contrast associated with noise or blockiness, which in pseudo-MOS data corresponds to low image quality, is not enhanced.

\section{Limitations and Future Work}

While MILO delivers strong performance in both image quality prediction and optimization tasks, it has limitations. The dataset used for training the \textit{Ensemble} components contains human scores collected in uncontrolled, online settings. As a result, the quality of the augmented training data used for our visual masking model is inherently limited by the noise and variability in the individual viewing conditions, such as distance to the display and ambient room lighting. 
The MILO visibility map is not calibrated using any threshold-based visibility data, unlike some existing metrics that attempt this calibration \cite{mantiuk2011hdr, mantiuk2021fovvideovdp}. However, accurately predicting per-pixel image differences as perceived by human observers remains an open challenge for any existing metric. 
Instead, in this paper, we focus on relative spatial differences in distortion visibility across varying image content, which provide meaningful information for both network training and test-time guidance in diffusion-based generative tasks.
In such generative scenarios, MILO cannot correct for failures of the underlying generative model; if the diffusion process produces semantically incorrect outputs, the metric has limited ability to guide recovery. %Finally, since ImageNet lacks human faces, performance on face-centric content is sometimes suboptimal due to insufficient representation during training.

%Future work will address these limitations by incorporating content-specific datasets, such as those containing human faces, to improve generalization across semantic categories. 
As future work, we plan to use generative models to augment data when explicit forward models are unavailable, such as for complex distortions in PIPAL. Lastly, we aim to extend MILO to a no-reference formulation by integrating content-aware perceptual priors.

\section{Conclusion}

We presented MILO, a fast and accurate multiscale metric for full-reference image quality assessment, trained by pseudo-MOS supervision without human annotations. Beyond prediction, MILO serves as an effective perceptual loss in both image and latent space. Applied at inference time to diffusion-based restoration, it enables efficient and perceptually guided optimization, improving quality in tasks like denoising, super-resolution, and face restoration. 
% MILO offers a compact and generalizable solution for FR-IQA and modern perceptual optimization pipelines.

\begin{acks}
We gratefully acknowledge the support of the Saarland/Intel Joint Program on the Future of Graphics and Media.
\end{acks}

\bibliographystyle{ACM-Reference-Format}
\bibliography{main}

% \newpage

% \input{sections/7_fig-only}

% \begin{figure}[htbp]
%     \centering
%     \includegraphics[width=0.8\linewidth]{figures/alpha_schedule_plot.png}
%     \caption{Comparison of cosine and linear scheduling for $\alpha$ over training epochs. Cosine scheduling increases slowly at first and accelerates in the middle, providing a smoother curriculum.}
%     \label{fig:alpha_schedule}
% \end{figure}

% \appendix

\end{document}